\documentclass{svproc}

\usepackage{amsmath}
\usepackage{amssymb}

\usepackage[table]{xcolor}
\usepackage{times,cite}
\usepackage{url}
\usepackage{graphicx}
\usepackage{listings}
\usepackage{color}

\definecolor{dkgreen}{rgb}{0,0.6,0}
\definecolor{gray}{rgb}{0.5,0.5,0.5}
\definecolor{mauve}{rgb}{0.58,0,0.82}

\begin{document}
\mainmatter              
\title{Fractals2019: Combinatorial Optimisation with \\ Dynamic Constraint Annealing}
%
\titlerunning{Fractals2019: Dynamic Constraint Annealing}  
%
\author{Mikhail Prokopenko$^{1,2}$ \and Peter Wang$^{2}$}
\authorrunning{Prokopenko and Wang} 
\institute{Complex Systems Research Group, Faculty of Engineering\\ 
			The University of Sydney, NSW 2006, Australia\\
\email{mikhail.prokopenko@sydney.edu.au}\\
\and
Data Mining, CSIRO Data61, PO Box 76, Epping, NSW 1710, Australia\\
}

\maketitle              

\begin{abstract}
Fractals2019 started as a new experimental entry in the RoboCup Soccer 2D Simulation League, based on Gliders2d code base, and advanced to become a RoboCup-2019 champion. We employ combinatorial optimisation methods, within the framework of Guided Self-Organisation, with the search guided by local constraints. We present examples of several tactical tasks based on the Gliders2d code (version v2), including the search for an optimal assignment of heterogeneous player types, as well as blocking behaviours, offside trap, and attacking formations. We propose a new method, \emph{Dynamic Constraint Annealing}, for solving dynamic constraint satisfaction problems, and apply it to optimise thermodynamic potential of collective behaviours, under dynamically induced constraints. 
\end{abstract}

\section{Introduction}
\vspace{-2mm}

The RoboCup Soccer 2D Simulation League provides a rich dynamic environment, facilitated by the RoboCup Soccer Simulator (RCSS), aimed to test advances in  decentralised collective behaviours of autonomous agents. The challenges include concurrent adversarial actions, computational nondeterminism, noise and latency in asynchronous perception and actuation, and limited processing time \cite{Stone99,Reis2001,Noda2003,Prokopenko03,Akiyama2008,gabel2008brainstormers,Bai2015-acm,RAM15,Prokopenko16}.  
The League progress has been supported by several important base code releases, covering both low-level skills and standardised world models of simulated agents \cite{LNAI99-simulator,Kok03robocup,HELIOSBase,marlik}. The release in 2010 of the base code of HELIOS team, \emph{agent2d-3.0.0}, later upgraded to agent2d-3.1.1, was particularly influential. By 2016, about 80\% of the teams adopted agent2d as their base code, including the champion team, Gliders2016 \cite{gliders2016tdp,Prokopenko16}, which also used fragments of MarliK source code \cite{marlik}, and by 2019 this fraction exceeded 90\%.

Gliders2016 was developed using Human-based Evolutionary Computation (HBEC) \cite{kosorukoff2001human,gliders2016tdp}.  
In 2018, we released the code base \emph{Gliders2d} \cite{G2d}, version v1, comprising six HBEC steps.  In this paper we present the second version of \emph{Gliders2d} (v2),  with six additional steps:
\begin{itemize}
\item Gliders2d-v2.1: blocking behaviour (disrupting the opponents in possession of the ball; based on simplified MarliK source code \cite{marlik}; \verb|bhv_basic_move.cpp|);
\item Gliders2d-v2.2: offside trap behaviour (\verb|bhv_basic_move.cpp|);
\item Gliders2d-v2.3: assignment of heterogeneous player types (\verb|sample_coach.cpp|); 
\item Gliders2d-v2.4: back action (allowing players to select actions which evaluate worse than the current action; \verb|action_chain_graph.cpp|);
\item Gliders2d-v2.5: tackle action (higher risk in defense); \verb|bhv_basic_move.cpp|);
\item Gliders2d-v2.6: wing attacking formation (\verb|strategy.cpp|).
\end{itemize}

Some of these steps are relatively simple, while some involve substantial code changes. Nevertheless, the performance improvements are tangible, and we will detail these in Section \ref{gliders2}. 
Gliders2d uses librcsc 4.1.0 \cite{HELIOSBase}. It is different from the competition branch (Gliders2012 --- Gliders2016), being a separate evolutionary branch, created over 2018--2019 to experiment with Fractals2019. We will exemplify how Guided Self-Organisation (GSO) allows us to optimise the performance, using the transition between v2.2 to v2.3 which improved an assignment of heterogeneous player types. 

Fractals2019 is a new team which is partially based on Gliders2d \cite{Fractals2019-TDP}, while retaining some elements of our previous champion team Gliders2016 \cite{gliders2016tdp,Prokopenko16}.  To a large extent, Fractals2019 is an experimental entry, motivated by a new set of aims.  Specifically, we redefined the fitness landscape used by optimisation in terms of dynamic constraints, rather than in terms of the performance metrics alone.  Our overall approach uses \emph{guided self-organisation} of tactical behaviour, shown here for the transition between v2.2 to v2.3, now extended by a thermodynamic characterisation of collective action, Section \ref{Methods}, with results described in Section \ref{results}.

\vspace{-3mm}
\section{Gliders2d: version v2}
\label{gliders2}
\vspace{-2mm}

Each HBEC solution can be seen as a ``genotype'', encoding the entire team behaviour in a set of ``design points'' \cite{Cioppa2007,G2d}, which may vary from a single parameter  (e.g., blocking depth), to an ordering of heterogeneous types with respect to some criterion (e.g., a list of integers representing players' roles), to complex multi-agent communication schemes \cite{Prokopenko16,Zuparic2017,gabelcommunication}.  

Each solution is typically evaluated against a specific opponent, over thousands of games, with the fitness function being the average goal difference, while the average points and standard error provide tie-breakers \cite{G2d}. In other words, a design point (possibly conditioned on the name of a specific opponent) is accepted only if it outperforms every single opponent in the pool of available opponents.  This strict acceptance criterion may be adjusted by assigning different weights to the available opponents, set in proportion to their respective strengths which are ``statically'' determined in advance. These weights can be based on the competition ranking of the previous year, or can be specified more precisely, using the points achieved in a multi-game round-robin tournament involving these opponents \cite{ProkopenkoWMBLC17}.  In this evolutionary/optimisation approach, the overall fitness is determined as a weighted average computed across the opponents in the pool. As a consequence, a design point may be accepted even it underperformed against a relatively weaker opponent (if it performed strongly against other, stronger, opponents). Rarely, a design point may even be accepted if it underperformed against the strongest opponent, provided that the other performances outweighed this loss on average. 

The weighted fitness function, accommodating relative strengths of the opponents, achieves several aims.  Firstly, it better accounts for the absence of transitivity in teams’ relative strengths \cite{ProkopenkoWMBLC17}. Secondly, it reduces the development and computation time required to ensure that any given design point outperforms all available opponents.  Thirdly, it produces a simple and more general-purpose code, with fewer conditional branches.   It is interesting to point out the similarity of the weighted fitness function with the concept of ``Nash averaging'' recently introduced by Balduzzi et al. \cite{Balduzzi2018}, who distinguished between two basic scenarios: agent-vs-agent and agent-vs-task.  Nash averaging was utilised in a ``dynamic'' setting, where at each round the strength of the evolving team (the ``agent'') itself contributed to the relative strengths of the opponents. Most recently this was carried out by DeepMind in the AlphaStar League, while evolving an agent excelling in the real-time strategy game StarCraft II.  The final AlphaStar agent comprised the elements of the Nash distribution of the league, capturing the most effective mixture of strategies discovered by the evolution guided by Nash average \cite{AlphaStar}.
In RoboCup Simulation League scenarios, this would mean that the weights of the opponents in a pool are re-evaluated in a round-robin tournament which includes the evolving solution (design point) itself, providing a better quality of the solutions --- but at a prohibitive computational cost (at this stage).  

A balanced pool of opponents (an ``ecosystem'') in which Gliders2d were evolved from v2.1 to v2.6, using a weighted fitness function, included four benchmark teams: Gliders2013 \cite{gliders2013tdp}, the 2018 world champion team, HELIOS2018 \cite{helios18}, the 2018 third-ranked team, MT2018 \cite{mt2018}, and the 2018 sixth-ranked team, YuShan2018 \cite{yushan2018}. For each sequential step and for the baseline Gliders2d-v1.6 (the latest step in version v1), 2000 games were played against these four benchmarks.  

Against Gliders2013, the goal difference achieved by Gliders2d-v2.6 improves from $-0.2$ to zero. Against YuShan2018, the goal difference improves from $-0.8$ to zero, achieving parity as well. Against MT2018, the goal difference improves from $-2.1$ to $-1.0$. Finally, against HELIOS2018, the goal difference improves from $-4.4$ to $-2.4$.  The latter case will be detailed specifically, for the transition between v2.2 and v2.3, when  the goal difference improved from $-4.2$ to $-3.0$, i.e., by more than a single goal, merely due to a better assignment of heterogeneous players. Table \ref{tel1} summarises the performance dynamics, including the overall goal difference and the standard error of the mean, for each of the match-ups. Adoption of the weighted fitness function ensures the progression at several steps: for example, from v1.6 to v2.1.  

\begin{table}[ht]
\vspace{-5mm}
\begin{center}
\begin{tabular}{|c|c|c|c|c|c|c|c|c|}
 \hline
 \scriptsize{ \ Version \ } & \multicolumn{2}{|c|}{\scriptsize{Gliders2013}} & \multicolumn{2}{|c|}{\scriptsize{YuShan2018}} & \multicolumn{2}{|c|}{\scriptsize{MT2018}} &  \multicolumn{2}{|c|}{\scriptsize{HELIOS2018}} \\ \hline
 \scriptsize{ \  \ } &  \scriptsize{ \ Goal diff. \ } &  \scriptsize{ \ Std. error  \ } & \scriptsize{ \ Goal diff. \ } & \scriptsize{ \ Std. error \ } & \scriptsize{ \ Goal diff. \ } & \scriptsize{ \ Std. error \ } &  \scriptsize{ \ Goal diff. \ } &  \scriptsize{ \ Std. error  \ } \\ \hline

  \scriptsize{ \ v1.6 \ } &  \scriptsize{ \ -0.242073 \ } &  \scriptsize{ \ 0.029355  \ } & \scriptsize{ \ -0.827273 \ } & \scriptsize{ \ 0.032821 \ } & \scriptsize{ \ -2.05796   \ } & \scriptsize{ \ 0.040560 \ } &  \scriptsize{ \ -4.43894  \ } &  \scriptsize{ \ 0.046226  \ } \\ \hline
	
	  \scriptsize{ \ v2.1 \ } &  \scriptsize{ \ -0.124561  \ } &  \scriptsize{ \ 0.028637  \ } & \scriptsize{ \ -0.499497  \ } & \scriptsize{ \ 0.030966 \ } & \scriptsize{ \ -1.74109   \ } & \scriptsize{ \ 0.038903 \ } &  \scriptsize{ \ -4.48316  \ } &  \scriptsize{ \ 0.046027  \ } \\ \hline

		  \scriptsize{ \ v2.2 \ } &  \scriptsize{ \ -0.128321  \ } &  \scriptsize{ \ 0.028243 \ } & \scriptsize{ \ -0.461384  \ } & \scriptsize{ \ 0.031010 \ } & \scriptsize{ \ -1.52841   \ } & \scriptsize{ \ 0.038033 \ } &  \scriptsize{ \ -4.17144   \ } &  \scriptsize{ \ 0.045290  \ } \\ \hline

			  \scriptsize{ \ v2.3 \ } &  \scriptsize{ \ -0.103206  \ } &  \scriptsize{ \ 0.029718  \ } & \scriptsize{ \ -0.422634  \ } & \scriptsize{ \ 0.031381 \ } & \scriptsize{ \ -1.33676   \ } & \scriptsize{ \ 0.036293 \ } &  \scriptsize{ \ -2.95471$^{\star}$  \ } &  \scriptsize{ \ 0.013678   \ } \\ \hline

				  \scriptsize{ \ v2.4 \ } &  \scriptsize{ \ -0.129081  \ } &  \scriptsize{ \ 0.029207  \ } & \scriptsize{ \ -0.190190  \ } & \scriptsize{ \ 0.029751 \ } & \scriptsize{ \ -1.27683   \ } & \scriptsize{ \ 0.034613 \ } &  \scriptsize{ \ -2.72030   \ } &  \scriptsize{ \ 0.036411  \ } \\ \hline

					  \scriptsize{ \ v2.5 \ } &  \scriptsize{ \ -0.138122  \ } &  \scriptsize{ \ 0.030924  \ } & \scriptsize{ \ -0.128192  \ } & \scriptsize{ \ 0.029237 \ } & \scriptsize{ \ -1.12130   \ } & \scriptsize{ \ 0.033411 \ } &  \scriptsize{ \ -2.52632   \ } &  \scriptsize{ \ 0.034567  \ } \\ \hline

						  \scriptsize{ \ v2.6 \ } &  \scriptsize{ \ -0.030105  \ } &  \scriptsize{ \ 0.028551  \ } & \scriptsize{ \ -0.016525  \ } & \scriptsize{ \ 0.033414 \ } & \scriptsize{ \ -1.01404   \ } & \scriptsize{ \ 0.035441 \ } &  \scriptsize{ \ -2.39538   \ } &  \scriptsize{ \ 0.036255  \ } \\ \hline
\end{tabular}
\end{center}
\caption{Performance evaluation of Gliders2d against benchmarks, over $\sim \! \! 2000$ games for each version of Gliders2d against the opponent. The match-up marked by ${\star}$ involved $\sim \! \!16000$ games (Section \ref{results}).}
\label{tel1}
\vspace{-9mm}
\end{table}

The released code, including the six sequential steps comprising version v2, is located at:

 \verb|http://www.prokopenko.net/gliders2d.html|.  

The differences can be easily identified by consecutively comparing the files indicated for each step across versions v2(n) and v2(n+1), for $n \geq 0$ (with v2.0 being v1.6). For example, differentiating the file \verb|bhv_basic_move.cpp| between v1.6 and v2.1 will show the changes implementing the blocking behaviour (with the acknowledged fragments of Marlik source code \cite{marlik}), while checking the differences in the same file between v2.1 and v2.2 will highlight the offside trap behaviour of Gliders2d. The offside trap is produced by collective motion of the defenders synchronised by their perception of relevant situational variables. The changes in the last three steps are self-explanatory, and also involve several variables describing situational patterns, forming design points evolved by HBEC or GSO.
The step from v2.2 to v2.3 deserves a more detailed analysis, being quite illustrative of the difference between HBEC and GSO approaches, and showing how self-organisation can be guided by thermodynamic properties of collective behaviour.

\vspace*{-3mm}
\section{Fractals2019: thermodynamically driven collective behaviours}
\label{Methods}
\vspace*{-2mm}

Guided Self-Organisation constrains self-organisation within a dynamical system to paths leading towards specific attractors or outcomes. It integrates two  techniques: (i) self-organising exploration of the search space, and (ii) traditional design following a ``blueprint'' \cite{pro09b,prokopenko2013guided}. GSO has been studied in several robotic scenarios, combining (i) universal objective functions, and (ii) task-dependent constraints on the system dynamics, often using generic information-theoretic or thermodynamic methods \cite{nehaniv2005evolutionary,prokopenko06-alife,sab06,der07,klyubin,Ay08,DerMartius11,prokopenko2013guided,Prokopenko-Einav,Hamann}. An example of combinatorial optimisation carried out in a noisy potential field using information-theoretic tools is described by Kim et al. \cite{Kim2011} in a study which complemented hill-climbing using information entropy. The class of simulated annealing algorithms is also characterised by a thermodynamic analogy \cite{Laarhoven}.

\vspace{-3mm}
\subsection{Problem formulation}
\vspace{-2mm}

In statistical-mechanical terms, each candidate solution (a design point) can be interpreted as a configuration of the suitably defined statistical system, so that the  fitness of the solution can be considered as the energy of the system in that particular configuration. Hence, one may in general represent an optimisation problem as a search for an equilibrium state of a system which optimises the appropriate thermodynamic potential (e.g., minimises the free energy).  

In developing Fractals2019 we combine (i) the standard (weighted) fitness function defined by the goal difference, which is inevitably measured only imprecisely, and (ii) local constraints which indirectly represent collective behaviour. The constraints are defined with respect to the elements of the design point under consideration, restricting the search-space. In HBEC approach, these constraints are specified before the search begins, by human designers. Importantly, under our GSO approach, these  constraints are not given in advance, but are  induced during the first phase of combinatorial optimisation which is driven by maximising the goal difference alone. Given the landscape of the fitness function, partially discovered during the first phase, the method identifies the regions around local maxima, and induces partial constraints that represent these local fitness sub-spaces. At the second phase of optimisation, we use the thermodynamic analogy and guide the search from the attained local optimum towards the constrained sub-space, shaped by the discovered constraints. 

For example, a design point can be specified as an assignment of heterogeneous player types to player roles, represented as an ordered list of integers, e.g., the assignment of \emph{agent2d-3.1.1}  sets $X = \{11, 2, 3, 10, 9, 6, 4, 5, 7, 8\}$, with function $\rho: \mathbb{N} \rightarrow \mathbb{N}$ defining the rank in this list, e.g., $\rho(11) = 1$. This list is used by the coach agent in assigning the strongest heterogeneous player type (defined according to some criterion, for instance, the fastest player type) to the first player from this list, i.e., player 11 (centreforward). The second best type is assigned to the second player on the list, i.e., to player 2 (left central defender), and so on, so that the weakest, slowest, type is assigned to player 8 (right midfielder). The goalkeeper, player 1, is assigned a type separately. A constraint can be specified as a preference over the ranks, e.g., $\rho(6) < \rho(10)$ would mean that player 6 should have preference in the assignment over player 10.  In general, ranking constraints have the form: $\rho(i) < \rho(j)$ for players $i$ and $j$. 

Finding the assignment which maximises the fitness function $f_X$ over all possible $10!$ candidate solutions $X \in \cal{X}$ in the search-space is the specific optimisation problem solved at the transition between v2.2 and v2.3. Even on a high-performance computing cluster with 100 two-minute games running in parallel, checking all 3,628,800 permutations over 1000 games each (to account for the score fluctuations due to the simulation noise), would require over 138 years of continuous computation. A hill-climbing optimisation method would cut this time considerably but is likely to find only inferior local optima. We advocate instead a GSO approach which combines elements of hill-climbing optimisation and dynamic constraint satisfaction problems, with insights from simulated annealing, informed by the thermodynamic analogy.

One may think of our assignment problem as a variant of the Traveling Salesman Problem (TSP), where each player needs to be assigned only one role, and the objective is to maximise the fitness (rather than minimise the cost).  The ``distances'' between the players (the nodes along a directed path) are not known, so we cannot simply aggregate the path segments into the total path length $d$. Instead, the approximate overall path distance is provided by the noisy fitness function (the average goal difference resulting from the assignment), and so the values $f_X$ and $f_Y$ of two neighbouring solutions $X$ and $Y$ can produce the difference $\delta = f_X - f_Y$. 

\vspace{-3mm}
\subsection{Phase 1: Hill-climbing and dynamically induced constraints}
\vspace{-2mm}

During the first phase, we use a  hill-climbing algorithm \cite{russell2016} based on a variant of insertion sort \cite{knuth1998}.  
Hill climbing  is  a greedy local search algorithm which always moves in the direction of increasing fitness $f$ by comparing with values of local neighbours, terminating when it reaches a solution without neighbours with a higher fitness. Defining a suitable neighbour function  makes a significant impact on the success of this algorithm. In TSP problems one often defines a simple $k$-node flip neighbour function where a sequence of $k$ nodes of one solution is flipped to produce a candidate.  For example, a 2-flip neighbour function applied to $X = \{11, 2, 3, 10, 9, 6, 4, 5, 7, 8\}$ at rank 4  produces a candidate $Y = \{11, 2, 3, 9, 10, 6, 4, 5, 7, 8\}$, with the ranks of players 10 and 9 reversed, not unlike a bubble sort algorithm that would compare $f_X$ and $f_Y$ before deciding if the flip should be accepted.  
Insertion sort is a more efficient sorting algorithm that produces a sorted list by finding a location for a given element within the current list.  Starting with the initial assignment, the insertion sort algorithm picks one element from the data (e.g., in the ranked order), and finds the optimal location for this element within the list by comparing the fitness values corresponding to different locations. Having inserted this element, the algorithm iterates to the next element, until all elements are properly inserted. At every location test, only a better candidate is accepted, following the hill climbing approach.  

The challenge is that the fitness function $f_Y$ must be computed for every location choice, that is, thousands of games need to be played for the assignment $Y$ against the benchmark. This is needed in order to reduce the effects of fluctuations --- an aspect particularly relevant when assigning heterogeneous types which are stochastically generated before each game. Nevertheless, the fitness remains noisy even for a large number of games, making hill climbing problematic.  

To address this challenge and reduce the overall number of tests, we make two extensions. Firstly, we use a heuristic: an assumption of ``iteration'' convexity, that is, we assume that the fitness landscape is convex along the insertion path of each element. Secondly, in addition to identifying the best location for an element, the algorithm checks the  neighbourhood of the identified location, and induces local constraints over the ranks, whenever possible. This approach is broadly in line with dynamic constraint satisfaction problem (DCSP) solvers in which the constraints are dynamically evolved while the CSP is being solved \cite{Verfaillie1994}. It differs from constraint recording techniques, in which newly learned constraints represent inconsistencies in the problem formulation, rather than the fitness landscape as such. 

Let us consider the iteration of inserting 11, traced in Table \ref{t1}. Initially the rank of the player is 1, i.e., $\rho(11) = 1$, and the corresponding fitness is $f = -4.17144$. Several locations are then iteratively tested, until a local maximum $f^* = -3.89289$ is found at rank $\rho(11) = 4$. The assumption of convexity along an iteration allows us to stop after the first maximum is identified during the iteration, that is, after checking the rank $\rho(11) = 5$ with the inferior fitness.  
At this stage we induce ranking constraints for 11: specifically, $\rho(10) < \rho(11)$ is induced by comparing tests 2 and 3, and $\rho(11) < \rho(9)$ is induced by comparing tests 3 and 4. Basically, the assignment differences between the tests are converted into constraints, to match the corresponding differences in fitness (if these differences are sufficiently large). For example, the difference between the tests 2 and 3 is the 2-flip at rank 3, picking 11 and 10, and since $f_2 < f_3$, it is induced that $\rho(10) < \rho(11)$. This preference impacts the collective behaviour of the team, resulting in the overall fitness (the thermodynamic potential), and so the fitness is used to induce the ranking.

Inducing such local constraints may appear redundant, as the  maximum identified in this example reflects them already.  However, not every iteration will improve on the current maximum, but may still identify local constraints that partially represent the structure of the search-space (see Section \ref{results} for more examples,  summarised in Table \ref{t2}). All these constraints will be used in the second phase, guiding a more refined search.

Our use of DCSP is motivated by the noisy fitness function. The fluctuations in the fitness function appear due to a dynamic and distributed RoboCup environment where the outcomes change from game to game --- and so estimating the fitness across multiple games forms a changing  problem environment. In general, problems tend to have structure, and the local constraints induced during the first phase partially discover this structure.

\begin{table}[ht]
\vspace{-3mm}
\begin{center}
\begin{tabular}{|c|c|c|c|c|}
 \hline
\scriptsize{ \ Test $i$ \ } &  \scriptsize{ \ Assignment \ }  & \scriptsize{ \ Goal difference $f_i$ \ } & \scriptsize{ \ Standard error \ } & \scriptsize{ \ Induced constraint \ } \\ \hline
\multicolumn{5}{|c|}{ \scriptsize{ \ Inserting 11 \ }} \\ \hline
\scriptsize{ \ 0 \ } &  \scriptsize{ \ 11 2 3 10 9 6 4 5 7 8 \ }  & \scriptsize{ \ -4.17144 \ } & \scriptsize{ \ 0.045290 \ } & \\ \hline
\scriptsize{ \ 1 \ } &  \scriptsize{ \ 2 11 3 10 9 6 4 5 7 8 \ }  & \scriptsize{ \ -4.04819 \ } & \scriptsize{ \ 0.065453 \ } & \\ \hline
\scriptsize{ \ 2 \ } &  \scriptsize{ \ 2 3 11 10 9 6 4 5 7 8 \ }  & \scriptsize{ \ -4.04100  \ } & \scriptsize{ \ 0.061556 \ } & \\ \hline
\scriptsize{ \ 3 \ } &  \scriptsize{ \ 2 3 10 11 9 6 4 5 7 8 \ }  & \scriptsize{ \ -3.89289$^*$ \ } & \scriptsize{ \ 0.061798 } \ & \scriptsize{\ $\rho(10) < \rho(11)$ \ } \\ \hline
\scriptsize{ \ 4 \ } &  \scriptsize{ \ 2 3 10 9 11 6 4 5 7 8 \ }  & \scriptsize{ \ -4.06928 \ } & \scriptsize{ \ 0.060939 \ } & \scriptsize{\ $\rho(11) < \rho(9)$ \ } \\ \hline
\end{tabular}
\end{center}
\caption{Hill climbing with insertion sort. A local maximum, $\rho(11) = 4$, is marked $*$. Two constraints are induced: $\rho(10) < \rho(11)$  and $\rho(11) < \rho(9)$.  Each test $i > 0$ involved $\sim \! \!1000$ games, test 0:  $\sim \! \!2000$ games.}
\label{t1}
\vspace{-5mm}
\end{table}

\vspace{-7mm}
\subsection{Phase 2: Constraint satisfaction via annealing}
\vspace{-1mm}

After carrying out iterations for all elements of the design point, and obtaining a local maximum $X$, as well as a set of local constraints, we begin the second phase guided by a thermodynamic characterisation of the fitness landscape. 
In a seminal work, \v{C}ern{\'y} \cite{cerny} argued that the analogy with thermodynamics offers a new insight into optimisation problems such as TSP: the length $d_X$ of a given trip, defined as a sequence/permutation of the nodes (in our terms, a given assignment of the player types), can be seen as the energy $E_X$ of the system in that particular configuration $X$. Crucially, ``simulating the transition to the equilibrium and decreasing the temperature, one can find smaller and smaller values of the mean energy of the system (= length of the trip)'' \cite{cerny}.  This argument provided strong motivation for simulated annealing algorithms \cite{Laarhoven,eglese}, where the probability of a candidate solution $X$ generated during an exploration of the search-space, while minimising the path length, is given by the Boltzmann-Gibbs distribution, for some temperature $T$ and normalisation constant $Z$:
\begin{equation}
P(X) = Z e^{-E_X / T} \ .
\end{equation}  
Given a current minimum $d_Y$, the chances of accepting a worse candidate $X$ with $d_X > d_Y$, are not zero, but depend on the (energy) difference $\delta = d_X - d_Y$. More precisely, a worse candidate is accepted with probability $e^{-\delta / T}$.  When the annealing temperature becomes sufficiently low, the acceptance probability for non-optimising solutions reduces as well. 

We use simulated annealing in guiding the exploration of the search-space. 
Once a locally optimal solution $X$ is obtained, we generate candidate solutions $Y$ in some proximity of the local optimum, but keeping closer to, and preferably within, the region of the search-space restricted by the discovered local constraints (``closer'' in terms of a simple neighbourhood function, e.g., insertion sort again).  The probability of accepting a worse candidate $Y$, with $d_X > d_Y$ is given by $e^{-\delta / T}$, for $\delta = d_X - d_Y$, with acceptable $Y$ replacing $X$ as the current best candidate. Thus, this algorithm shares \v{C}ern{\'y}'s insight quoted earlier, in exploring the search-space thermodynamically, i.e., along the landscape of a thermodynamic potential, guided by local constraints towards an optimum (equilibrium) state.     
In the beginning of the search, when the temperature $T$ is higher, acceptable solutions can be found deeper within the constrained sub-space, but when the search cools down, candidate solutions may sit very close to the currently identified optimum.  
Section \ref{results} demonstrates the second search phase for our main example, see Table \ref{t3}.

We refer to our second phase as \emph{constraint satisfaction via annealing}. It differs from the ``constraint annealing'' technique \cite{Kropaczek} which interprets constraints as objective functions and applies standard simulated annealing to the redefined problem. The main overall difference, of course,  lies in our use of dynamically induced constraints, and so we refer to the entire optimisation method introduced in this paper as \emph{Dynamic Constraint Annealing}.

In general, the objective function and, thus the dynamic local constraints, may be chosen to represent different aspects of the problem structure: maximisation of spatiotemporal coordination \cite{prokopenko06-alife,sab06}, maximisation of information flows \cite{cliff2013towards,CliffLWWOP17}, maximisation of thermodynamic efficiency which contrasts the gain in the uncertainty reduction with the required work \cite{Crosato2018}, etc.  

\vspace{-3mm}
\section{Results}
\label{results}
\vspace{-2mm}

In this section we develop our leading example, detailing the two optimisation phases.  Let us continue with the hill-climbing phase: having inserted element 11, we iterate insertion sort algorithm for other elements, while applying the convexity heuristic and avoiding testing the assignments which have been checked earlier, as shown in Table \ref{t2}. 

\begin{table}
\begin{center}
\begin{tabular}{|c|c|c|c|c|}
 \hline
\scriptsize{ \ Test $i$ \ } &  \scriptsize{ \ Assignment $X_i$ \ }  & \scriptsize{ \ Goal difference $f_i$ \ } & \scriptsize{ \ Standard error \ } & \scriptsize{ \ Induced constraint \ } \\ \hline
\scriptsize{ \ 3 \ } &  \scriptsize{ \ 2 3 10 11 9 6 4 5 7 8 \ }  & \scriptsize{ \ -3.89289$^*$ \ } & \scriptsize{ \ 0.061798 } \ & \scriptsize{\ $\rho(10) < \rho(11)$ \ } \\ \hline
\hline

\multicolumn{5}{|c|}{ \scriptsize{ \ Inserting 2 \ }} \\ \hline
\scriptsize{ \ 5 \ } &  \scriptsize{ \ 3 2 10 11 9 6 4 5 7 8 \ }  & \scriptsize{ \ -3.96985 \ } & \scriptsize{ \ 0.064817 \ } & \scriptsize{\ $\rho(2) < \rho(3)$ \ }  \\ \hline

\multicolumn{5}{|c|}{ \scriptsize{ \ Inserting 3 \ }} \\ \hline
\scriptsize{ \ 6 \ } &  \scriptsize{ \ 2 10 3 11 9 6 4 5 7 8 \ }  & \scriptsize{ \ -4.24600 \ } & \scriptsize{ \ 0.066165 \ } & \scriptsize{\ $\rho(3) < \rho(10)$ \ } \\ \hline

\multicolumn{5}{|c|}{ \scriptsize{ \ Inserting 10 \ }} \\ \hline
\scriptsize{ \ 7 \ } &  \scriptsize{ \ 10 2 3 11 9 6 4 5 7 8	\ }  & \scriptsize{ \ -4.23547 \ } & \scriptsize{ \ 0.063354 \ } & \scriptsize{\ $[\rho(2) < \rho(10)]$ \ } \\ \hline

\multicolumn{5}{|c|}{ \scriptsize{ \ Inserting 9 \ }} \\ \hline
\scriptsize{ \ 8 \ } &  \scriptsize{ \ 9 2 3 10 11 6 4 5 7 8	\ }  & \scriptsize{ \ -4.14766 \ } & \scriptsize{ \ 0.064273 \ } &   \\ \hline

\scriptsize{ \ 9 \ } &  \scriptsize{ \ 2 9 3 10 11 6 4 5 7 8	\ }  & \scriptsize{ \ -4.23695 \ } & \scriptsize{ \ 0.065580 \ } &   \\ \hline 

\scriptsize{ \ 10 \ } &  \scriptsize{ \ 2 3 9 10 11 6 4 5 7 8	\ }  & \scriptsize{ \ -4.08133  \ } & \scriptsize{ \ 0.063088 \ } &   \\ \hline \hline
 
\scriptsize{ \ 4 \ } &  \scriptsize{ \ 2 3 10 9 11 6 4 5 7 8 \ }  & \scriptsize{ \ -4.06928 \ } & \scriptsize{ \ 0.060939 \ } & \\ \hline

\scriptsize{ \ 3 \ } &  \scriptsize{ \ 2 3 10 11 9 6 4 5 7 8 \ }  & \scriptsize{ \ -3.89289$^*$ \ } & \scriptsize{ \ 0.061798 } \ & \scriptsize{\ $\rho(11) < \rho(9)$ \ } \\ \hline \hline 

\scriptsize{ \ 11 \ } &  \scriptsize{ \ 2 3 10 11 6 9 4 5 7 8	\ }  & \scriptsize{ \ -3.93493  \ } & \scriptsize{ \ 0.063252 \ } &  \scriptsize{\ $[\rho(6) < \rho(9)]$ \ }  \\ \hline

\multicolumn{5}{|c|}{ \scriptsize{ \ Inserting 6 \ }} \\ \hline
\scriptsize{ \ 12 \ } &  \scriptsize{ \ 6 2 3 10 11 9 4 5 7 8	\ }  & \scriptsize{ \ -4.03722  \ } & \scriptsize{ \ 0.065353 \ } &   \\ \hline

\scriptsize{ \ 13 \ } &  \scriptsize{ \ 2 6 3 10 11 9 4 5 7 8	\ }  & \scriptsize{ \ -4.14874  \ } & \scriptsize{ \ 0.061345 \ } &   \\ \hline

\scriptsize{ \ 14 \ } &  \scriptsize{ \ 2 3 6 10 11 9 4 5 7 8	\ }  & \scriptsize{ \ -3.98894  \ } & \scriptsize{ \ 0.065420 \ } &  \scriptsize{\ $\rho(3) < \rho(6)$ \ }  \\ \hline 

\scriptsize{ \ 15 \ } &  \scriptsize{ \ 2 3 10 6 11 9 4 5 7 8	\ }  & \scriptsize{ \ -4.07251  \ } & \scriptsize{ \ 0.062459 \ } &  \scriptsize{\ $\rho(6) < \rho(10)$ \ }  \\ \hline 
 
\multicolumn{5}{|c|}{ \scriptsize{ \ Inserting 4 \ }} \\ \hline
\scriptsize{ \ 16 \ } &  \scriptsize{ \ 4 2 3 10 11 9 6 5 7 8	\ }  & \scriptsize{ \ -3.79618$^*$ \ } & \scriptsize{ \ 0.060502 \ } &   \\ \hline

\scriptsize{ \ 17 \ } &  \scriptsize{ \ 2 4 3 10 11 9 6 5 7 8	\ }  & \scriptsize{ \ -3.72417$^*$ \ } & \scriptsize{ \ 0.059761 \ } &   \\ \hline

\scriptsize{ \ 18 \ } &  \scriptsize{ \ 2 3 4 10 11 9 6 5 7 8	\ }  & \scriptsize{ \ -3.69539$^*$ \ } & \scriptsize{ \ 0.058036 \ } &  \scriptsize{\ $[\rho(3) < \rho(4)]$ \ } \\ \hline

\scriptsize{ \ 19 \ } &  \scriptsize{ \ 2 3 10 4 11 9 6 5 7 8	\ }  & \scriptsize{ \ -3.80542 \ } & \scriptsize{ \ 0.060862 \ } &  \scriptsize{\ $\rho(4) < \rho(10)$ \ } \\ \hline

\multicolumn{5}{|c|}{ \scriptsize{ \ Inserting 5 \ }} \\ \hline
\scriptsize{ \ 20 \ } &  \scriptsize{ \ 5 2 3 4 10 11 9 6 7 8	\ }  & \scriptsize{ \ -3.22312$^*$ \ } & \scriptsize{ \ 0.055354 \ } &   \\ \hline

\scriptsize{ \ 21 \ } &  \scriptsize{ \ 2 5 3 4 10 11 9 6 7 8	\ }  & \scriptsize{ \ -3.19860$^*$ \ } & \scriptsize{ \ 0.052708 \ } &   \\ \hline

\scriptsize{ \ 22 \ } &  \scriptsize{ \ 2 3 5 4 10 11 9 6 7 8	\ }  & \scriptsize{ \ -3.15800$^*$ \ } & \scriptsize{ \ 0.056941 \ } & \scriptsize{\ $[\rho(3) < \rho(5)]$ \ } \\ \hline

\scriptsize{ \ 23 \ } &  \scriptsize{ \ 2 3 4 5 10 11 9 6 7 8	\ }  & \scriptsize{ \ -3.28815 \ } & \scriptsize{ \ 0.055315 \ } &  \scriptsize{\ $\rho(5) < \rho(4)$ \ } \\ \hline

\multicolumn{5}{|c|}{ \scriptsize{ \ Inserting 7 \ }} \\ \hline

\scriptsize{ \ 24 \ } &  \scriptsize{ \ 7 2 3 5 4 10 11 9 6 8	\ }  & \scriptsize{ \ -3.44869 \ } & \scriptsize{ \ 0.059632 \ } &  \\ \hline

\scriptsize{ \ 25 \ } &  \scriptsize{ \ 2 7 3 5 4 10 11 9 6 8	\ }  & \scriptsize{ \ -3.41141 \ } & \scriptsize{ \ 0.059249 \ } &  \\ \hline

\scriptsize{ \ 26 \ } &  \scriptsize{ \ 2 3 7 5 4 10 11 9 6 8	\ }  & \scriptsize{ \ -3.35772 \ } & \scriptsize{ \ 0.059014 \ } &  \\ \hline

\scriptsize{ \ 27 \ } &  \scriptsize{ \ 2 3 5 7 4 10 11 9 6 8	\ }  & \scriptsize{ \ -3.31795 \ } & \scriptsize{ \ 0.057105 \ } &  \\ \hline

\scriptsize{ \ 28 \ } &  \scriptsize{ \ 2 3 5 4 7 10 11 9 6 8	\ }  & \scriptsize{ \ -3.18036 \ } & \scriptsize{ \ 0.057603 \ } & \scriptsize{\ $\rho(4) < \rho(7)$ \ } \\ \hline

\scriptsize{ \ 29 \ } &  \scriptsize{ \ 2 3 5 4 10 7 11 9 6 8	\ }  & \scriptsize{ \ -3.34604 \ } & \scriptsize{ \ 0.057504 \ } & \scriptsize{\ $\rho(7) < \rho(10)$ \ } \\ \hline

\multicolumn{5}{|c|}{ \scriptsize{ \ Inserting 8 \ }} \\ \hline

\scriptsize{ \ 30 \ } &  \scriptsize{ \ 8 2 3 5 4 10 11 9 6 7	\ }  & \scriptsize{ \ -3.39980 \ } & \scriptsize{ \ 0.058509 \ } &  \\ \hline

\scriptsize{ \ 31 \ } &  \scriptsize{ \ 2 8 3 5 4 10 11 9 6 7	\ }  & \scriptsize{ \ -3.41106 \ } & \scriptsize{ \ 0.058709 \ } &  \\ \hline

\scriptsize{ \ 32 \ } &  \scriptsize{ \ 2 3 8 5 4 10 11 9 6 7	\ }  & \scriptsize{ \ -3.31891 \ } & \scriptsize{ \ 0.056437 \ } &  \\ \hline

\scriptsize{ \ 33 \ } &  \scriptsize{ \ 2 3 5 8 4 10 11 9 6 7	\ }  & \scriptsize{ \ -3.30924 \ } & \scriptsize{ \ 0.055972 \ } &  \\ \hline

\scriptsize{ \ 34 \ } &  \scriptsize{ \ 2 3 5 4 8 10 11 9 6 7	\ }  & \scriptsize{ \ -3.12261$^*$ \ } & \scriptsize{ \ 0.057369 \ } & \scriptsize{\ $\rho(4) < \rho(8)$ \ } \\ \hline

\scriptsize{ \ 35 \ } &  \scriptsize{ \ 2 3 5 4 10 8 11 9 6 7	\ }  & \scriptsize{ \ -3.19157 \ } & \scriptsize{ \ 0.056433 \ } & \scriptsize{\ $\rho(8) < \rho(10)$ \ } \\ \hline

\end{tabular}
\end{center}
\caption{Hill climbing with insertion sort: continuing after first four tests which inserted 11 at rank 4 (Table \ref{t1}). The current maxima are marked with $*$. Tests 3 and 4 are shown repeatedly, to clarify the comparisons which induced constraints. When the difference between fitness values is smaller than the standard error, a possible constraint, shown within $[ \ ]$, is not induced. Each test is carried out over $\sim \! \!1000$ games.}
\label{t2}
\end{table}

\begin{figure}[ht]
\centering
\includegraphics[width=8.0cm]{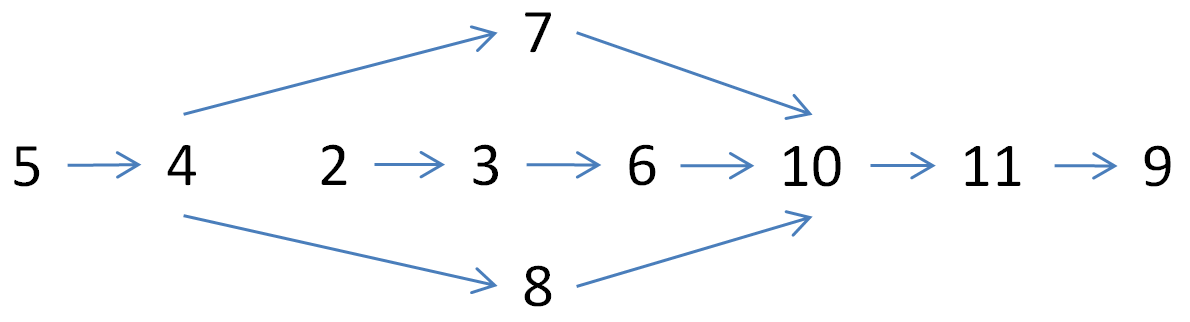} 
\caption{A partial ordering underlying an optimal assignment implied by induced local constraints. Ranking preference $\rho(i) < \rho(j)$ is depicted as $i \rightarrow j $.} 
\label{fig1}
\vspace{-3mm}
\end{figure}

The dynamically induced local constraints construct a partial ordering underlying an optimal assignment, shown in Figure \ref{fig1}.  It is evident that the local maximum attained by test 34 does not satisfy all these local constraints, highlighting the need for a second, constraint satisfaction, phase. The algorithm continues with the best solution $X_{34} = \{2, 3, 5, 4, 8, 10, 11, 9, 6, 7\}$, re-evaluated after 16000 games in order to ensure a higher precision, attaining $f_{34} = -3.14496$, and then generates the candidates closer to and within the constrained sub-space, as shown in  Table \ref{t3}, for a specified number of tests, e.g., 10 additional tests. The maximum attained at the end of this phase is given by $X_{44} = \{5, 4, 2, 3, 7, 6, 8, 10, 11, 9\}$ with $f_{44} = -2.95471$, reducing the goal deficit by further 0.2.  This solution fully satisfies the local constraints.

Not surprisingly, this assignment which was optimised  against the world champion of 2018, is assortative: it prefers to place the fastest players in defence (defenders 5, 4, 3, 2, with the wing defenders 5 and 4 assigned the best types), followed by midfielders (7, 6, 8), and leaving the weakest types for forwards (10, 11, 9). To re-iterate, this is carried out at the transition step from Gliders2d-v2.2 to v2.3, which is still a relatively weak team overall. In general, such an optimisation should be repeated after each tactical improvement, adjusting the collective behaviour of heterogeneous players to the latest tactics. In fact, every time when the optimal assignment changes in response to newest tactics, it signifies an important step in the development of collective behaviour. Our winning entry, Fractals2019, maintained a more balanced assignment than Gliders2d-v2.6, disassortatively mixing the roles across types to fit a more attacking style ---  the final optimisation was carried out in the last days before the championship. The Dynamic Constraint Annealing method described in this study takes about three days of computation on a high-performance cluster with 100 parallel games in the RCSS synchronisation mode (about 2 minutes per game), with the first and second phase taking roughly 12 hours ($\sim \! \!35000$ games) and 54 hours  ($\sim \! \!160000$ games) respectively. And so such a re-optimisation becomes feasible.

Optimisation of other design points can also be carried out with Dynamic Constraint Annealing, given a suitably chosen discretisation of situational variables, so that relevant local constraints can be induced during the iterative hill-climbing phase.

\begin{table}
\vspace{-4mm}
\begin{center}
\begin{tabular}{|c|c|c|c|c|c|}
 \hline
\scriptsize{ \ Test $i$ \ } &  \scriptsize{ \ Assignment $X_i$ \ }  & \scriptsize{ \ Goal difference $f_i$ \ } & \scriptsize{ \ Standard error \ } & \scriptsize{ \ Annealing temp. $T$ \ } & \scriptsize{ \ Acceptance prob. $P$ \ } \\ \hline

\scriptsize{ \ 34 \ } &  \scriptsize{ \ 2 3 5 4 8 10 11 9 6 7	\ }  & \scriptsize{ \ -3.14496$^*$ \ } & \scriptsize{ \ 0.014086 \ } & \scriptsize{ \  \ } & \scriptsize{ \ \ } \\ \hline
 \hline
\scriptsize{ \ 36 \ } &  \scriptsize{ \ 2 5 3 4 8 10 11 9 6 7	\ }  & \scriptsize{ \ -3.12690$^*$ \ } & \scriptsize{ \ 0.013852 \ } & \scriptsize{ \ 0.10 \ } & \scriptsize{ \ 1 \ } \\ \hline
\scriptsize{ \ 37 \ } &  \scriptsize{ \ 2 5 3 4 8 7 10 11 9 6	\ }  & \scriptsize{ \ -3.11121$^*$ \ } & \scriptsize{ \ 0.013912 \ } & \scriptsize{ \ 0.09 \ } & \scriptsize{ \ 1 \ } \\ \hline
\scriptsize{ \ 38 \ } &  \scriptsize{ \ 2 5 3 4 7 8 10 11 9 6	\ }  & \scriptsize{ \ -3.05126$^*$ \ } & \scriptsize{ \ 0.013869 \ } & \scriptsize{ \ 0.08 \ } & \scriptsize{ \ 1 \ } \\ \hline
\scriptsize{ \ 39 \ } &  \scriptsize{ \ 2 5 3 4 7 8 6 10 11 9	\ }  & \scriptsize{ \ -3.05799$^a$ \ } & \scriptsize{ \ 0.013767 \ } & \scriptsize{ \ 0.07 \ } & \scriptsize{ \ 0.90833 \ } \\ \hline
\scriptsize{ \ 40 \ } &  \scriptsize{ \ 2 5 3 4 7 6 8 10 11 9	\ }  & \scriptsize{ \ -3.04399$^*$ \ } & \scriptsize{ \ 0.013761 \ } & \scriptsize{ \ 0.06 \ } & \scriptsize{ \ 1 \ } \\ \hline
\scriptsize{ \ 41 \ } &  \scriptsize{ \ 5 2 3 4 7 6 8 10 11 9	\ }  & \scriptsize{ \ -3.11263$^r$ \ } & \scriptsize{ \ 0.014044 \ } & \scriptsize{ \ 0.05 \ } & \scriptsize{ \ 0.31854 \ } \\ \hline
\scriptsize{ \ 42 \ } &  \scriptsize{ \ 2 5 4 3 7 6 8 10 11 9	\ }  & \scriptsize{ \ -3.00621$^*$ \ } & \scriptsize{ \ 0.013789 \ } & \scriptsize{ \ 0.04 \ } & \scriptsize{ \ 1 \ } \\ \hline
\scriptsize{ \ 43 \ } &  \scriptsize{ \ 5 2 4 3 7 6 8 10 11 9	\ }  & \scriptsize{ \ -2.98825$^*$ \ } & \scriptsize{ \ 0.013473 \ } & \scriptsize{ \ 0.03 \ } & \scriptsize{ \ 1 \ } \\ \hline
\scriptsize{ \ 44 \ } &  \scriptsize{ \ 5 4 2 3 7 6 8 10 11 9	\ }  & \scriptsize{ \ -2.95471$^*$ \ } & \scriptsize{ \ 0.013678 \ } & \scriptsize{ \ 0.02 \ } & \scriptsize{ \ 1 \ } \\ \hline
\scriptsize{ \ 45 \ } &  \scriptsize{ \ 5 4 2 3 6 7 8 10 11 9	\ }  & \scriptsize{ \ -2.96470$^r$ \ } & \scriptsize{ \ 0.013397 \ } & \scriptsize{ \ 0.01 \ } & \scriptsize{ \ 0.36825 \ } \\ \hline

\end{tabular}
\end{center}
\caption{Constraint satisfaction via annealing: continuing after the first phase (Table \ref{t2}) for 10 additional tests. The current maxima are marked with $*$, worse but accepted solutions -- by $a$, and worse but rejected solutions -- by $r$. Each test is carried out over $\sim \! \!16000$ games.}
\label{t3}
\vspace{-6mm}
\end{table}

\vspace*{-6mm} 
\section{Conclusions}
\vspace*{-2mm}

Team Fractals2019 is based on recently released Gliders2d code base \cite{G2d}. The second version of Gliders2d is described and traced in this study against a pool of benchmark opponents, using a fitness function weighted by relative strengths of the benchmarks.  We followed the methodology of Guided Self-Organisation in  optimising for strong team performance, within a noisy fitness landscape affected by fluctuations inherent in the nondeterministic RoboCup simulation environment. This is achieved by employing a thermodynamic analogy in characterising the global objective function, as well as local constraints which are dynamically discovered during the combinatorial optimisation. We proposed a new method, \emph{Dynamic Constraint Annealing}, which improves on greedy search such as hill-climbing techniques, by (i) dynamically inducing local constraints, and (ii) guiding the constraint satisfaction phase by a simulated annealing carried out along a trajectory approaching or within the constrained sub-space. We illustrated this method for a specific problem, an optimal assignment of heterogeneous player types to player roles, interpreted as a variant of the Traveling Salesman Problem (TSP), demonstrating a tangible increase of fitness over a series of tests. This technique may be generally applicable to combinatorial optimisation and constraint satisfaction problems in changing distributed environments.

Fractals2019 allowed us to verify the applicability of the GSO approach to combinatorial optimisation problems within a challenging  environment of RoboCup simulation. 
RoboCup-2019 competition  included 15 teams from 7 countries: Australia, Brazil, China, Germany, Iran, Japan, and Portugal. Fractals2019 played 22 games during several rounds, winning 16 times, with the total score of 59:10, or 2.68 : 0.45 on average. The final game against team HELIOS2019 (Japan) went into the extra time, and ended with Fractals2019 winning 1:0.  Post-tournament we carried out a 16000-game  experiment between the two finalists, using their released binaries \cite{Fractals2019-TDP,HELIOS2019-TDP}, with the champion team, Fractals2019, outperforming the runner-up by 0.22 goals (the average score 0:79:0.57) with 0.009 of standard error.

\textbf{Acknowledgments}. We thank HELIOS team for their excellent code base of \emph{agent2d}, as well as several members of Gliders team contributing during 2012--2016: David Budden, Oliver Cliff, Victor Jauregui and Oliver Obst.


\begin{thebibliography}{10}
\providecommand{\url}[1]{\texttt{#1}}
\providecommand{\urlprefix}{URL }
\providecommand{\doi}[1]{https://doi.org/#1}

\bibitem{HELIOSBase}
Akiyama, H., Nakashima, T.: {HELIOS Base: An Open Source Package for the
  RoboCup Soccer 2D Simulation}. In: Behnke, S., Veloso, M., Visser, A., Xiong,
  R. (eds.) RoboCup 2013: Robot World Cup XVII. pp. 528--535. Springer, Berlin,
  Heidelberg (2014)

\bibitem{HELIOS2019-TDP}
Akiyama, H., Nakashima, T., Fukushima, T., Suzuki, Y., Ohori, A.: Helios2019:
  Team description paper. In: RoboCup 2019 Symposium and Competitions, Sydney,
  Australia (2019)

\bibitem{Akiyama2008}
Akiyama, H., Noda, I.: Multi-agent positioning mechanism in the dynamic
  environment. In: Visser, U., Ribeiro, F., Ohashi, T., Dellaert, F. (eds.)
  {RoboCup 2007: Robot Soccer World Cup XI}, pp. 377--384. Springer, Berlin,
  Heidelberg (2008)

\bibitem{Ay08}
Ay, N., Bertschinger, N., Der, R., Guttler, F., Olbrich, E.: Predictive
  information and explorative behavior of autonomous robots. Eur. Phys. J. B
  \textbf{63},  329--339(11) (2008)

\bibitem{Bai2015-acm}
Bai, A., Wu, F., Chen, X.: Online planning for large {Markov} decision
  processes with hierarchical decomposition. ACM Trans. Intell. Syst. Technol.
  \textbf{6}(4),  45:1--45:28 (2015)

\bibitem{Balduzzi2018}
Balduzzi, D., Tuyls, K., Perolat, J., Graepel, T.: Re-evaluating evaluation.
  In: Proc. of the 32nd Int'l Conf. on Neural Information Processing Systems.
  pp. 3272--3283. USA (2018)

\bibitem{RAM15}
Budden, D.M., Wang, P., Obst, O., Prokopenko, M.: {RoboCup} simulation leagues:
  Enabling replicable and robust investigation of complex robotic systems. IEEE
  Trans. Robot. Autom.  \textbf{22}(3),  140--146 (2015)

\bibitem{cerny}
{\v{C}}ern{\'y}, V.: Thermodynamical approach to the traveling salesman
  problem: An efficient simulation algorithm. J. Optimiz. Theory App.
  \textbf{45}(1),  41--51 (Jan 1985)

\bibitem{yushan2018}
Cheng, Z., Xie, N., Sun, C., Tan, C., Zhang, K., Wang, L., Zhang, G., She, X.,
  Zheng, X.: {YuShan2018 Team Description Paper for RoboCup2018}. In: RoboCup
  2018 Symposium and Competitions, Montreal, Canada (2018)

\bibitem{Cioppa2007}
Cioppa, T.M., Lucas, T.W.: Efficient nearly orthogonal and space-filling
  {Latin} hypercubes. Technometrics  \textbf{49}(1),  45--55 (2007)

\bibitem{cliff2013towards}
Cliff, O.M., Lizier, J., Wang, R., Wang, P., Obst, O., Prokopenko, M.: Towards
  quantifying interaction networks in a football match. In: Behnke, S., Veloso,
  M., Visser, A., Xiong, R. (eds.) RoboCup 2013: Robot Soccer World Cup XVII.
  pp. 1--12. Springer (2013)

\bibitem{CliffLWWOP17}
Cliff, O., Lizier, J., Wang, X.R., Wang, P., Obst, O., Prokopenko, M.:
  Quantifying long-range interactions and coherent structure in multi-agent
  dynamics. Artif. Life  \textbf{23}(1),  34--57 (2017)

\bibitem{Crosato2018}
Crosato, E., Spinney, R.E., Nigmatullin, R., Lizier, J.T., Prokopenko, M.:
  Thermodynamics and computation during collective motion near criticality.
  Phys. Rev. E  \textbf{97},  012120 (2018)

\bibitem{AlphaStar}
DeepMind: {AlphaStar: Mastering the Real-Time Strategy Game StarCraft II}
  (2019),
  https://deepmind.com/blog/article/alphastar-mastering-real-time-strategy-game-starcraft-ii

\bibitem{DerMartius11}
Der, R., Martius, G.: {T}he {P}layful {M}achine -- {T}heoretical Foundation and
  Practical Realization of Self-Organizing Robots. Springer (2012)

\bibitem{eglese}
Eglese, R.W.: {Simulated annealing: A tool for operational research}. Eur. J.
  Oper. Res.  \textbf{46}(3),  271--281 (1990)

\bibitem{gabelcommunication}
Gabel, T., Kl{\"o}ppner, P., Godehardt, E., Tharwat, A.: Communication in
  soccer simulation: On the use of wiretapping opponent teams. In: RoboCup
  2018: Robot Soccer World Cup XXII. Springer (2018)

\bibitem{Hamann}
Hamann, H., Khaluf, Y., Botev, J., Soorati, M.D., Ferrante, E., Kosak, O.,
  Montanier, J.M., Mostaghim, S., Redpath, R., Timmis, J., Veenstra, F., Wahby,
  M., Zamuda, A.: Hybrid societies: Challenges and perspectives in the design
  of collective behavior in self-organizing systems. Front. Robot. AI
  \textbf{3}, ~14 (2016)

\bibitem{Kim2011}
Kim, P., Nakamura, S., Kurabayashi, D.: Hill-climbing for a noisy potential
  field using information entropy. Paladyn  \textbf{2}(2),  94--99 (2011)

\bibitem{klyubin}
Klyubin, A., Polani, D., Nehaniv, C.: Representations of space and time in the
  maximization of information flow in the perception-action loop. Neural
  Comput.  \textbf{19}(9),  2387--2432 (2007)

\bibitem{knuth1998}
Knuth, D.E.: The art of computer programming, vol.~3. Addison-Wesley (1997)

\bibitem{Kok03robocup}
Kok, J.R., Vlassis, N., Groen, F.: {U}v{A} {T}rilearn 2003 team description.
  In: Polani, D., Browning, B., Bonarini, A., Yoshida, K. (eds.) Proc. CD
  RoboCup 2003. Springer (2003)

\bibitem{kosorukoff2001human}
Kosorukoff, A.: Human based genetic algorithm. In: Systems, Man, and
  Cybernetics, 2001 IEEE International Conference on. vol.~5, pp. 3464--3469.
  IEEE (2001)

\bibitem{Kropaczek}
Kropaczek, D.J., Walden, R.: Constraint annealing method for solution of
  multiconstrained nuclear fuel cycle optimization problems. Nucl. Sci. Eng.
  \textbf{193}(5),  506--522 (2019)

\bibitem{Laarhoven}
Laarhoven, P.J.M., Aarts, E.H.L. (eds.): Simulated Annealing: Theory and
  Applications. Kluwer Academic Publishers, Norwell, MA, USA (1987)

\bibitem{der07}
Martius, G., Herrmann, M., Der, R.: Guided self-organisation for autonomous
  robot development. In: {Almeida e Costa}, F., Rocha, L., Costa, E., Harvey,
  I., Coutinho, A. (eds.) Advances in Artificial Life: 9th European Conference
  on Artificial Life (ECAL-2007), Lisbon, Portugal. Lecture Notes in Artificial
  Intelligence, vol.~4648, pp. 766--775. Springer (2007)

\bibitem{helios18}
Nakashima, T., Akiyama, H., Suzuki, Y., Ohori, A., Fukushima, T.: {HELIOS2018:
  Team Description Paper}. In: RoboCup 2018 Symposium and Competitions,
  Montreal, Canada (2018)

\bibitem{nehaniv2005evolutionary}
Nehaniv, C., Polani, D., Olsson, L., Klyubin, A.: Evolutionary
  information-theoretic foundations of sensory ecology: Channels of
  organism-specific meaningful information. Modeling Biology: Structures,
  Behaviour, Evolution pp. 9--11 (2005)

\bibitem{Noda2003}
Noda, I., Stone, P.: {The RoboCup Soccer Server and CMUnited Clients:
  Implemented Infrastructure for MAS Research}. Auton. Agents Multi-Agent Syst.
   \textbf{7}(1--2),  101--120 (2003)

\bibitem{sab06}
Prokopenko, M., Gerasimov, V., Tanev, I.: Evolving spatiotemporal coordination
  in a modular robotic system. In: Nolfi, S., Baldassarre, G., Calabretta, R.,
  Hallam, J.C.T., Marocco, D., Meyer, J.A., Miglino, O., Parisi, D. (eds.) From
  Animals to Animats 9: 9th International Conference on the Simulation of
  Adaptive Behavior (SAB 2006), Rome, Italy, September 25-29, 2006. Lecture
  Notes in Computer Science, vol.~4095, pp. 558--569. Springer (2006)

\bibitem{prokopenko06-alife}
Prokopenko, M., Gerasimov, V., Tanev, I.: Measuring spatiotemporal coordination
  in a modular robotic system. In: Rocha, L., Yaeger, L., Bedau, M., Floreano,
  D., Goldstone, R., Vespignani, A. (eds.) Artificial Life X: Proceedings of
  The 10th International Conference on the Simulation and Synthesis of Living
  Systems. pp. 185--191. Bloomington IN, USA (2006)

\bibitem{gliders2013tdp}
Prokopenko, M., Obst, O., Wang, P., Budden, D., Cliff, O.M.: {Gliders2013:
  Tactical analysis with information dynamics}. In: RoboCup 2013 Symposium and
  Competitions, Eindhoven, The Netherlands (2013)

\bibitem{gliders2016tdp}
Prokopenko, M., Wang, P., Obst, O., Jaurgeui, V.: {Gliders2016: Integrating
  multi-agent approaches to tactical diversity}. In: RoboCup 2016 Symposium and
  Competitions, Leipzig, Germany (2016)

\bibitem{pro09b}
Prokopenko, M.: Guided self-organization. HFSP Journal  \textbf{3}(5),
  287--289 (2009)

\bibitem{prokopenko2013guided}
Prokopenko, M.: Guided self-organization: Inception, vol.~9. Springer (2013)

\bibitem{Prokopenko-Einav}
Prokopenko, M., Einav, I.: Information thermodynamics of near-equilibrium
  computation. Phys. Rev. E  \textbf{91},  062143 (2015)

\bibitem{Prokopenko03}
Prokopenko, M., Wang, P.: Evaluating team performance at the edge of chaos. In:
  Polani, D., Browning, B., Bonarini, A., Yoshida, K. (eds.) RoboCup 2003:
  Robot Soccer World Cup VII. Lecture Notes in Computer Science, vol.~3020, pp.
  89--101. Springer (2004)

\bibitem{Prokopenko16}
Prokopenko, M., Wang, P.: {Disruptive Innovations in RoboCup 2D Soccer
  Simulation League: From Cyberoos'98 to Gliders2016}. In: Behnke, S., Sheh,
  R., Sariel, S., Lee, D. (eds.) {RoboCup 2016: Robot World Cup {XX}}. LNCS,
  vol.~9776, pp. 529--541. Springer (2017)

\bibitem{Fractals2019-TDP}
Prokopenko, M., Wang, P.: Fractals2019: Guiding self-organisation of
  intelligent agents. In: RoboCup 2019 Symposium and Competitions, Sydney,
  Australia (2019)

\bibitem{G2d}
Prokopenko, M., Wang, P.: {Gliders2d: Source Code Base for RoboCup 2D Soccer
  Simulation League}. In: RoboCup 2019: Robot World Cup XXIII, Springer (2020,
  to appear), also: arxiv.org/abs/1812.10202

\bibitem{ProkopenkoWMBLC17}
Prokopenko, M., Wang, P., Marian, S., Bai, A., Li, X., Chen, X.: {RoboCup 2D
  Soccer Simulation League}: Evaluation challenges. In: Akiyama, H., Obst, O.,
  Sammut, C., Tonidandel, F. (eds.) RoboCup 2017: Robot World Cup {XXI}
  [Nagoya, Japan, July 27-31, 2017]. Lecture Notes in Computer Science, vol.
  11175, pp. 325--337. Springer (2017)

\bibitem{Reis2001}
Reis, L.P., Lau, N., Oliveira, E.: Situation based strategic positioning for
  coordinating a team of homogeneous agents. In: Balancing Reactivity and
  Social Deliberation in Multi-Agent Systems, From RoboCup to Real-World
  Applications. pp. 175--197. Springer (2001)

\bibitem{gabel2008brainstormers}
Riedmiller, M., Gabel, T., Trost, F., Schwegmann, T.: Brainstormers 2d -- team
  description 2008. In: RoboCup2008 (2008)

\bibitem{russell2016}
Russell, S.J., Norvig, P.: Artificial intelligence: a modern approach. New
  Jersey: Prentice Hall (2003)

\bibitem{LNAI99-simulator}
Stone, P., Riley, P., Veloso, M.: The {CMU}nited-99 champion simulator team.
  In: Veloso, M., Pagello, E., Kitano, H. (eds.) {R}obo{C}up-99: Robot Soccer
  World Cup {III}, Lecture Notes in Artificial Intelligence, vol.~1856, pp.
  35--48. Springer, Berlin (2000)

\bibitem{Stone99}
Stone, P., Veloso, M.: Task decomposition, dynamic role assignment, and
  low-bandwidth communication for real-time strategic teamwork. Artif. Intell.
  \textbf{110}(2),  241--273 (1999)

\bibitem{marlik}
Tavafi, A., Nozari, N., Vatani, R., Yousefi, M.R., Rahmatinia, S., Pirdir, P.:
  {MarliK 2012 Soccer 2D Simulation Team Description Paper}. In: RoboCup 2012
  Symposium and Competitions, Mexico City, Mexico (2012)

\bibitem{Verfaillie1994}
Verfaillie, G., Schiex, T.: Solution reuse in dynamic constraint satisfaction
  problems. In: Proceedings of the Twelfth National Conference on Artificial
  Intelligence (Vol. 1). pp. 307--312. AAAI'94, American Association for
  Artificial Intelligence, Menlo Park, CA, USA (1994)

\bibitem{mt2018}
Yang, Z., Liu, Z., Wang, X., Dong, N., Hu, X., Li, J., Chen, S., Lyu, G.:
  {MT2018: Team Description Paper}. In: RoboCup 2018 Symposium and
  Competitions, Montreal, Canada (2018)

\bibitem{Zuparic2017}
Zuparic, M., Jauregui, V., Prokopenko, M., Yue, Y.: Quantifying the impact of
  communication on performance in multi-agent teams. Artif. Life Robot.
  \textbf{22}(3),  357--373 (2017)

\end{thebibliography}

\end{document}